\documentclass[runningheads]{llncs}
\usepackage{graphicx}
\usepackage{comment}
\usepackage{amsmath,amssymb} 
\usepackage{color}

\usepackage{booktabs}
\usepackage{siunitx} 	
\definecolor{gray_ours}{rgb}{0.6, 0.6, 0.6}

\usepackage{hyperref} 	
\usepackage{subcaption}
\begin{document}
\pagestyle{headings}
\mainmatter

\title{A high fidelity synthetic face framework for computer vision} 


%
\author{Tadas Baltru\v{s}aitis, Erroll Wood, Virginia Estellers, Charlie Hewitt, Sebastian Dziadzio, Marek Kowalski, Matthew Johnson, Thomas J. Cashman, and Jamie Shotton}

\institute{Microsoft Mixed Reality \& AI Labs, Cambridge, UK\\
\{tabaltru, erwood, viestell, v-charhe, sedziadz, makowals, matjoh, tcashman,jamiesho\}@microsoft.com}
%
\authorrunning{Baltru\v{s}aitis et al.}
%
\maketitle

\begin{abstract}

Analysis of faces is one of the core applications of computer vision, with tasks ranging from landmark alignment, head pose estimation, expression recognition, and face recognition among others.
However, building reliable methods requires time-consuming data collection and often even more time-consuming manual anotation, which can be unreliable.
In our work we propose synthesizing such facial data, including ground truth annotations that would be almost impossible to acquire through manual annotation at the consistency and scale possible through use of synthetic data.
We use a parametric face model together with hand crafted assets which enable us to generate training data with unprecedented quality and diversity (varying shape, texture, expression, pose, lighting, and hair).



\end{abstract}



\section{Synthetic Face framework}

In this technical report we present our synthetic face generation pipeline.
We describe how each of the assets is generated, sampled, and represented for downstream machine learning tasks.

Our synthetic face model is made up of a learned 3D Face Model (Section \ref{sec:3dmodel}), textures based on scan data (Section \ref{sec:texture}), and a selection of hand crafted hair assets (Sections \ref{sec:hair}). 
We describe the constituent parts of our model and how it is rendered in the following sections.

\subsection{3D Face Model}
\label{sec:3dmodel}
The surface geometry of our face model is parametrized by an articulated control mesh $M$ with 7127 vertices $V$, a fixed topology of 6908 quad faces, and a skeleton $S$ with 4 joints controlling the pose of the neck, jaw, and each eye. During render time our face model uses subdivision surfaces to describes the resulting surface geometry $\mathcal{M}$.
An approximation to the smooth face surface $\mathcal{M}$ is obtained by three levels of Catmull-Clark subdivsion \cite{catmull1978recursively} applied to $M$.

\begin{figure}[ht]
\begin{subfigure}{.24\textwidth}
  \centering
  \includegraphics[width=.8\linewidth]{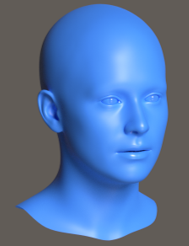}
  \caption{Template face}
  \label{fig:template_face}
\end{subfigure}
\begin{subfigure}{.24\textwidth}
  \centering
  \includegraphics[width=.8\linewidth]{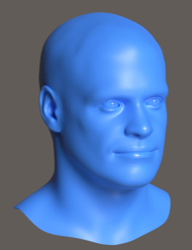}
  \caption{Identity mesh}
  \label{fig:apply_identity}
\end{subfigure}
\begin{subfigure}{.24\textwidth}
  \centering
  \includegraphics[width=.8\linewidth]{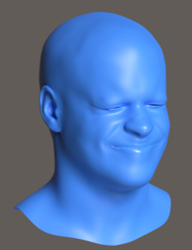}
  \caption{Unposed mesh $\bar{V}$}
  \label{fig:apply_expression}
\end{subfigure}
\begin{subfigure}{.24\textwidth}
  \centering
  \includegraphics[width=.8\linewidth]{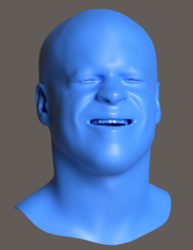}
  \caption{Posed mesh $V$}
  \label{fig:apply_pose}
\end{subfigure}
\caption{Illustration of the decomposition of our face generating model into identity, expression, and pose transforms: \ref{fig:apply_identity} shows the effects of applying the identity transform on the template face \ref{fig:template_face}, \ref{fig:apply_expression} shows the effect of applying the expression transform to \ref{fig:apply_identity}, and \ref{fig:apply_pose} shows the effects of applying the pose transform to \ref{fig:apply_expression}.}
\label{fig:parametric_model}
\end{figure}

The topology of $M$ is carefully designed to represent expressive human faces accurately, with higher resolution in areas subject to larger deformations (eyes, lips) when face identity or expression changes. While the mesh topology is fixed, its geometry is a function of model parameters $\theta= (\alpha, \beta, \gamma)$: first identity and expression parameters are used to generate a mesh $\bar{V}(\alpha, \beta)$ in canonical pose and then pose parameters $\gamma$ create a mesh in the desired pose. Figure \ref{fig:parametric_model} illustrates this decomposition.

The pose parameters $\gamma$ control the position of the mesh vertices through linear-blend skinning on the skeleton $S$.
In our model, the skinning weights have been manually designed and the rigid transform associated with each bone $T_i = [R_i| t_i]$ is a combination of a rotation $R_i(\gamma_i)$ parametrized by Euler angles and a translation vector $t_i = t^0_i + a_i \cdot \alpha$ that is a linear function of the identity parameters and the position of the bone $t^0_{i}$ in the template skeleton. The pose transform thus takes a mesh in canonical pose $\bar{V}$ and computes the vertices of the posed mesh as $V = \mathcal{T}(\alpha, \gamma, \bar{V})$.


The unposed mesh $\bar{V}$ is  the result of applying identity and expression deformations to a template face $\bar{V}_{0}$. We use 53 identity and 51 expression blendshapes to parametrize deformations as follows:
\begin{align}
\bar{V}(\alpha, \beta)= \bar{V_0} + \sum_{i=1}^{m} \alpha_i \phi_i + \sum_{i=1}^{m} \beta_i \varphi_i,
\end{align}
where the identity blendshapes $\phi_1, \ldots, \phi_m$ are learned from data and the template face $\bar{V}_{0}$ and expression blendshapes $\varphi_1, \ldots, \varphi_m$ are manually designed. To simplify notation, we define a single transform $F(\theta, \Phi) = \mathcal{T}(\alpha, \gamma, \bar{V}(\alpha, \beta))$ that goes from model parameters $\theta$ to vertices $V$.

To learn the identity blendshapes, we use a training dataset of high-quality scans of 101 different individuals with a neutral expression that have been manually registered\footnote{\url{https://www.russian3dscanner.com/}} to the topology of $M$. Once registered, the vertices of all the scans $V^{1}, \ldots, V^N$ are in correspondence with the vertices of $M$ and we can formulate the problem as finding the identity blendshapes that best explain our training data under the mesh generating function $F(\theta, \Phi)$. We measure how well a generated mesh $x$ explains each training sample $V^k$ with the loss function $\ell(V^{k}, x)$ and learn the identity basis by solving the minimization problem
\begin{align}
\min_{\Phi} & \sum_{k=1}^{N} \ell(V_{k}, \min_{\theta^k} F(\theta^k, \Phi))
\end{align}
To avoid the nested minimization, we lift the problem and optimize jointly over the model parameters associated to each mesh and the identity blendshapes:
\begin{align}
\min_{\theta^1, \ldots, \theta^{N}, \Phi} & \sum_{k=1}^{N} \ell(V_k, F(\theta^k, \Phi) ).
\end{align}
The loss function is the sum of data and regularization terms. The data terms measure the distance between the vertices and normals of the target mesh $V^k$ and the generated sample $F(\theta^k, \Phi)$ and the regularization terms constrain the model parameters to their support, solve the ambiguity inherent to scaling blendshapes or their coefficients, encourage spatially smooth blendshapes, and penalize the appearance of mesh artefacts like degenerate faces and self-intersections.

In particular, we use 4th-order polynomial barrier functions to constrain the expression coefficients into the unit interval and the pose parameters to angle limits that result in joint rotations that are anatomically possible (e.g. the neck cannot turn $180\deg$), least-squares penalties on the identity coefficients and blendshapes to solve the scale ambiguity, the norm of the mesh Laplacian to measure the smoothness of the blendshapes, and standard mesh metrics like edge length or curvature to prevent degenerate meshes.

Our learning strategy is similar to Li et al. \cite{FLAME:2017}, where instead of alternating the minimization with respect to the identity blendshapes (solved with PCA by Li et al. \cite{FLAME:2017}) and the model parameters, we lift the optimization problem and optimize the loss jointly with respect to identity blendshapes and model parameters. In our implementation, we also make use of automatic differentiation and solve the minimization problem with a gradient-based descent method (Adam) in TensorFlow. Our training data is of higher quality, additionally some of the scans are manually cleaned by artists in order to remove hair, hair coverings, and noise from the scanning process. You can see examples of the scans in Figure \ref{fig:cleaning}.

Our eye model is based on SynthesEyes model~\cite{Wood2015} and is made of two spheres. The first sphere represents the sclera (white of the eye) and is flattened at the end to represent the iris and the pupil.
The second sphere represents the corneal bulge, and is transparent, refractive (n=1.376), and partially reﬂective.
The eyelids are shrinkwrapped to the sphere of the sclera to avoid a gap between the geometries.

\textbf{Sampling} To sample from our face geometry model we train a Gaussian Mixture Model (GMM) on the parameters $\alpha_1, \ldots, \alpha_k$ of the model fit to our training data.
We sample face shape from the GMM by randomly selecting a mixture and sampling with variance $\sigma=0.8$

\textbf{Representation} We use the identity 53 basis parameters as a representation of face geometry.

\subsection{Expression library}

\begin{figure}[t]
    \centering
    \includegraphics[width=\textwidth]{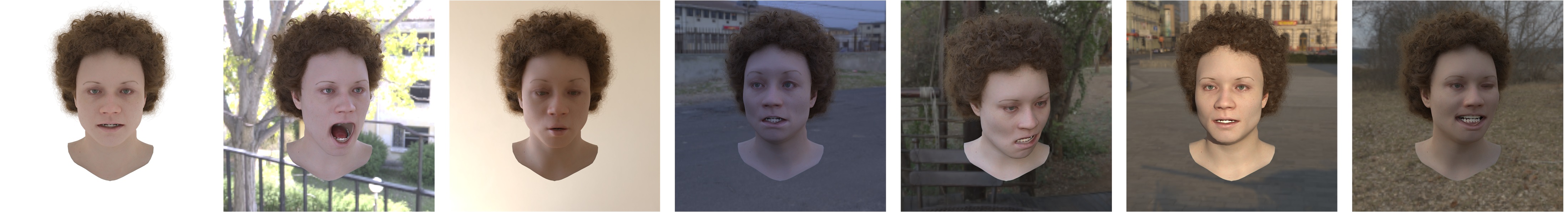}
    \caption{Sampling different expressions (pose and linear blendshapes) from our expression library on the same identity.}
    \label{fig:exp-sampling}
\end{figure}

As an expression library we fit a 3D face model to annotated 2D landmarks on images of faces.
The images are semi-automatically annotated using a facial landmark detector and manual inspection and correction.
This leads to a library of ~20k expressions.

\textbf{Sampling} To sample expression we randomly pick an expression from our library.
To sample pose (head, neck, and gaze) we draw from a normal distribution within physically possible ranges.
Further, when gaze is sampled we perform (with $50\%$ chance) blendshape correction to raise eyelids when looking up and lower them when looking down.
You can see different expressions sampled from the same subject in Figure \ref{fig:exp-sampling}.

\textbf{Representation} The expression is represented as 51 expression parameters (blendshapes) and 15 pose parameters.

\subsection{Texture}
\label{sec:texture}

\begin{figure}[t]
    \centering
    \includegraphics[width=\textwidth]{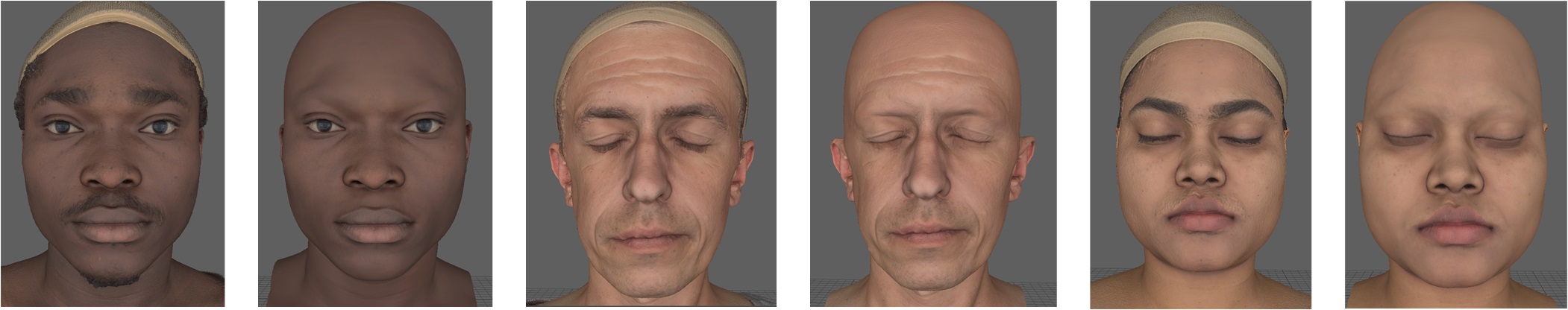}
    \caption{Sample scans with textures from our dataset, with corresponding raw and cleaned versions. Note the removal of hair, hair nets, and scanning artifacts.}
    \label{fig:cleaning}
\end{figure}

To represent the diffuse skin color, we use the aligned textures of scan data described in the previous section. However, instead of building a parametric model from them \cite{Wood2016,Guller2017}, we instead use the textures directly for added fidelity, especially for close-up or high resolution renders.
However, just applying the diffuse texture from the scan would lead to eyebrows, eyelashes, facial hair, hair nets and head hair being \emph{baked} into the texture.
To avoid this we manually clean a subset of 200 textures to remove such artefacts, both from the texture and the scan (see Figure \ref{fig:cleaning} for example of cleaned scans).

\begin{figure}[t]
    \centering
    \begin{subfigure}{0.32\textwidth}
        \includegraphics[width=\textwidth]{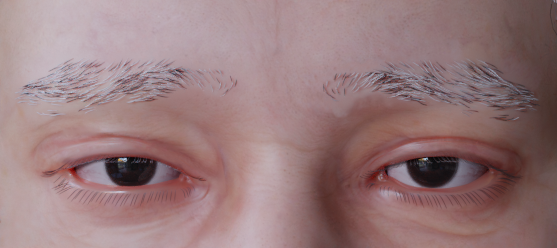}
    \end{subfigure}
    \begin{subfigure}{0.32\textwidth}
        \includegraphics[width=\textwidth]{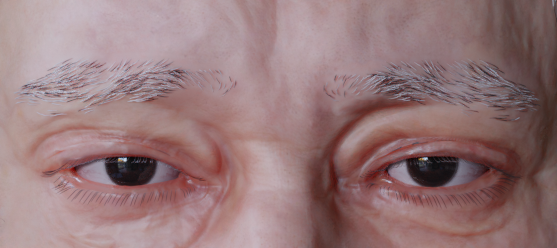}
    \end{subfigure}
    \begin{subfigure}{0.32\textwidth}
        \includegraphics[width=\textwidth]{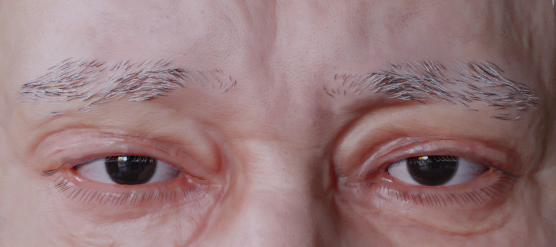}
    \end{subfigure}
    \caption{Using our meso and micro displacement maps in image rendering. Left to right: only using diffuse skin color, addition of meso displacement, addition of micro displacement. Note the added skin detail, especially for skin with wrinkles.}
\label{fig:displacement}
\end{figure}

As our face model is a fairly low resolution one, it leads to a smooth appearance of the skin (especially at glancing angles). To create a more realistic appearance we construct two displacement maps to model \emph{meso} and \emph{micro} displacement of the skin.
We construct the meso displacement map, by computing the displacement between the retopologized scan and the cleaned raw scan.
To model the micro displacement we compute a \emph{pore} map from the diffuse textue by applying a Laplacian of a Gaussian, which acts as a \emph{blob} detector, and can act as a micro displacent map. An example of effect of displacement maps can be seen in Figure \ref{fig:displacement}.

\begin{figure}[t]
    \centering
    \includegraphics[width=\textwidth]{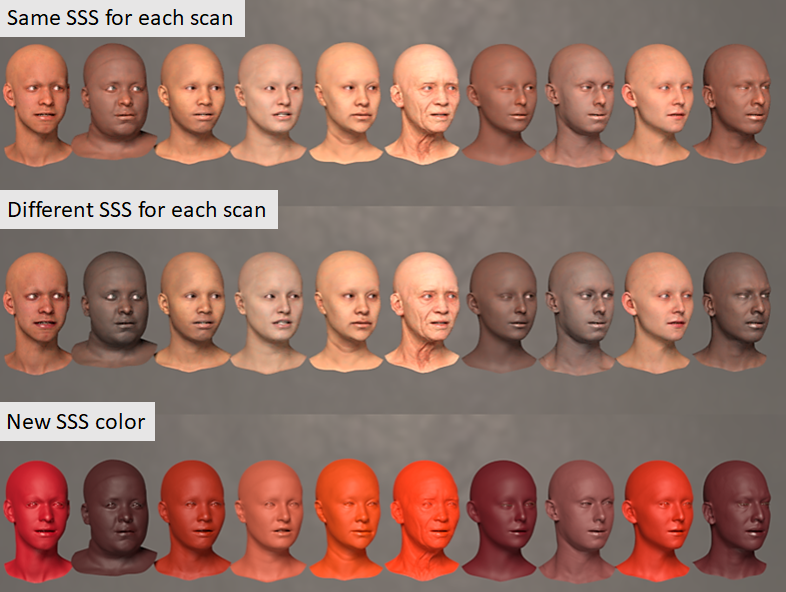}
    \caption{Using a different subsurface scaterring color for each sampled texture. Note how certain skin tones appear more realistic and less red.}
    \label{fig:sss}
\end{figure}

For added realism we also model skin specularity and skin subsurface scattering through manually created roughness and subsurface scattering maps (same maps used for every model).
We also find that using the same subsurface color for each sampled texture leads to less realistic results (see Figure \ref{fig:sss}), we therefore use the mean diffuse skin color to guide the subsurface color.

We combine the diffuse, roughness, displacement, and subsurface maps using a physically based shader \cite{Burley2012} as implemented by Blender's Cycles renderer\footnote{\url{https://www.blender.org/}}.

\textbf{Sampling}
During scene creation we randomly sample one of the 171 available diffuse textures together with two correponding displacement maps.
We further sample one of the manually created eye (out of five available colors) mouth, teeth and tongue textures.

\textbf{Representation}
We train a Variational Autoencoder (VAE) on the albedo texture data and use it's latent space as a low-dimensional representation of the texture data that can be easily consumed by machine learning methods. As the face region of the texture is the most salient one, we crop all the textures used in VAE training to contain that region only. The VAE is trained with a peceptual loss \cite{perceptual_loss} and mean squared error on the texture data as well as a KL loss on the latent space. The encoder and decoder consist of 6 convolutional layers each, the latent space has 50 dimensions.

\subsection{Hair}
\label{sec:hair}

\begin{figure}[t]
    \centering
    \begin{subfigure}{0.41\textwidth}
    \includegraphics[width=\textwidth]{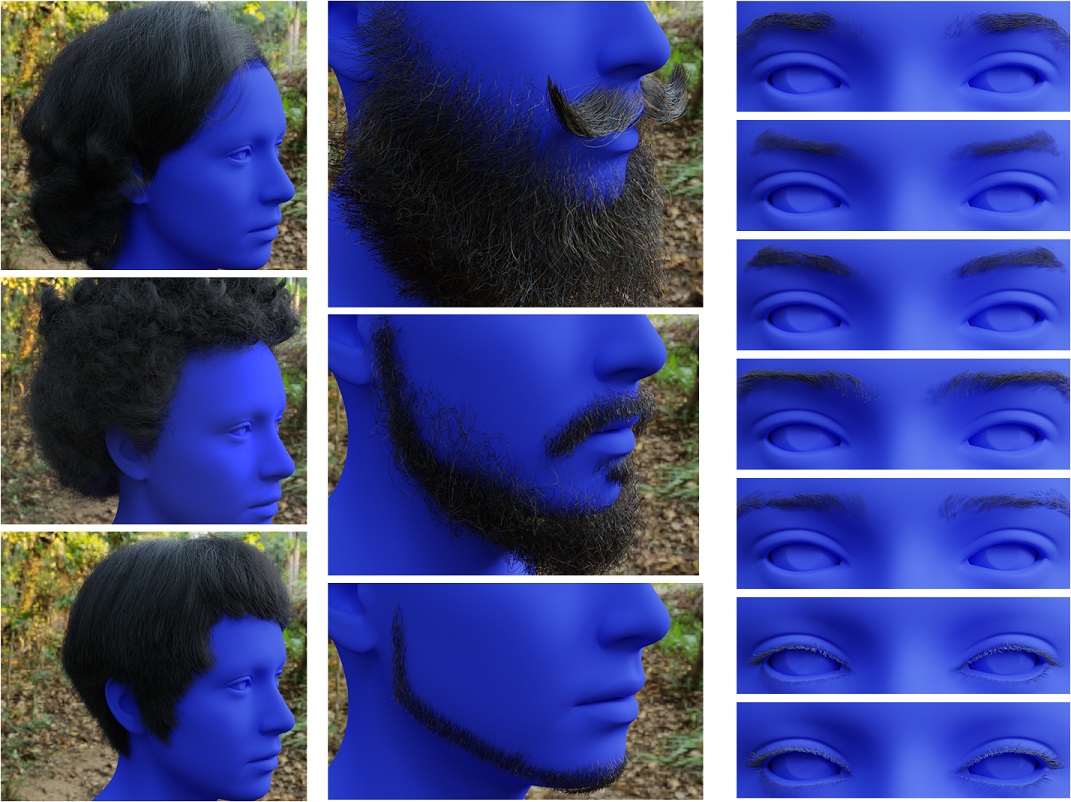}
    \caption{}
    \label{fig:hair_library}
    \end{subfigure}
    \begin{subfigure}{0.5\textwidth}
        \includegraphics[width=\textwidth]{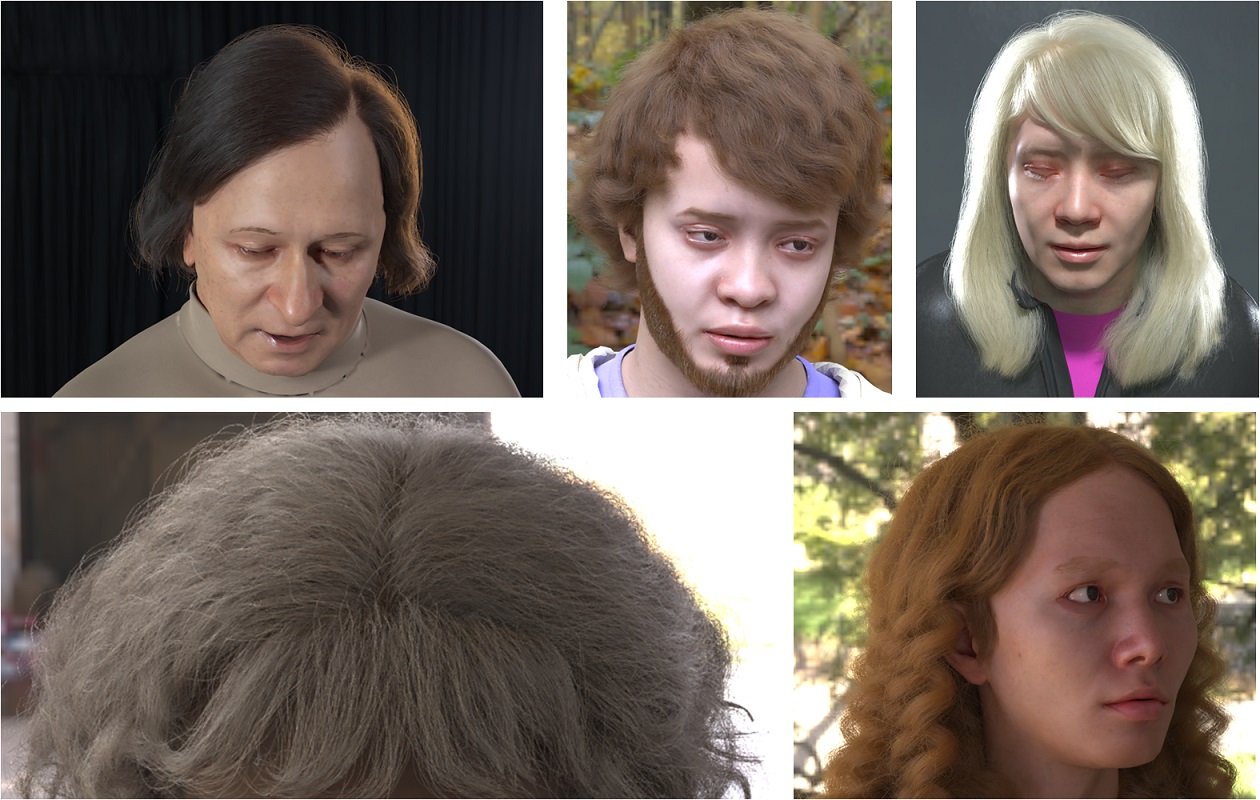}
        \caption{}
        \label{fig:hair_sampling}
        \end{subfigure}
    \caption{Examples of our hair synthetics}
\label{fig:all_hair}
\end{figure}

All hair in our synthetic face model is represented by 3D hair strands (for each individual hair).
This is in contrast to previous approaches where hair is baked into texture \cite{Guller2017,Wood2016}.
Such hair representation allows for more realistic representation of varying hair styles under different illuminations (see Figure \ref{fig:hair_sampling}).

Our hair library contains a broad selection of hair styles: 136 scalp; 50 eyebrows; 71 beards; and 3 eyelashes.
The grooms are created on the template head and deformed accordingly as the head shape changes (using Blender's particle hair system).
Some sample grooms from our hair library can be seen in Figure \ref{fig:hair_library}.

The hair color is represented as three scalars - proportion of melanin, pheomelanin, and proportion of gray hairs.
The strand based hair is shaded using a physically based hair shader~\cite{Chiang2016}.

\textbf{Sampling} We randomly sample from our manually created library of hair grooms.
Further, all of the grooms are flipped along the x-axis to double the size of the hair library.
To sample the hair color we use the world-wide statistics from Lozano et al.~\cite{Lozano2017}. Each of the hair grooms on the face is assigned the same color (e.g. brown eyebrows, hair, and beard). See Figure \ref{fig:hair_sampling}, for examples of hair colors sampled in our model.

\textbf{Representation} As each hair groom has a different number of hairs and number of segments for each hair, representing it is non-trivial.
We use a representation of hair based on a combination of volumetric flow direction and occupancy based paramterization, similar to Saito et al. ~\cite{Saito2018} and 2D UV image based parametrization.
We encode each hair style into: two UV maps (length and density) and a single volume map (flow direction).
This allows us to create a dimensionality reduced representation of each hair groom using principal component analysis.
Such a representation allows for easier use of the hair grooms in machine learning approaches, while still retaining the information required to reconstruct the hair.
See Figure

\begin{figure}[t]
    \centering
    \begin{subfigure}{0.47\textwidth}
    \includegraphics[width=\textwidth]{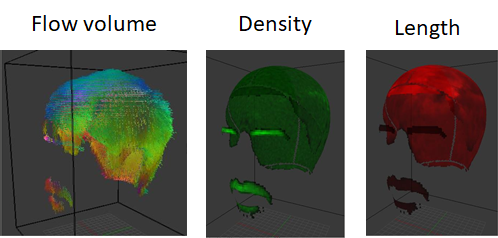}
    \caption{Visualization of flow volume, density and length parametrization of a hairstyle.}
    \label{fig:hair_library}
    \end{subfigure}
    \begin{subfigure}{0.35\textwidth}
        \includegraphics[width=\textwidth]{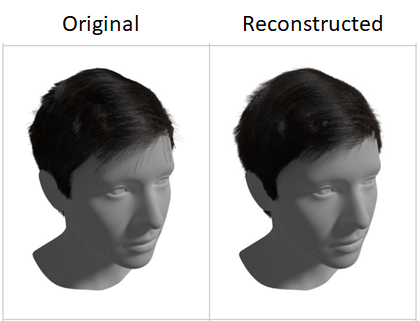}
        \caption{Original and reconstructed hairstyle after parametrization}
        \label{fig:hair_sampling}
        \end{subfigure}
    \caption{Hair parametrization and reconstruction.}
\label{fig:hair_reco}
\end{figure}





\subsection{Illumination}

Similar to Wood et al.~\cite{Wood2016}, to illuminate our face model, we use image-based lighting, a technique where high dynamic range (HDR) panoramic images are used to provide light in a scene \cite{Debevec2002}. This works by photographically capturing omni-directional light information, storing it in a
texture, and then projecting it onto a sphere around the object.
When a ray hits that texture during rendering, it takes that texture’s pixel value as light intensity.
In addition to providing illumination, using HDR images allows us to simulate backgrounds that are consistent with the illumination, without having to explicitly model background scenes.

\textbf{Sampling} We randomly choose and rotate one of 324 freely available HDR environment images \footnote{\url{https://hdrihaven.com/}, accessed February 2020} to simulate a range of different lighting conditions.

\textbf{Representation}
To create a low-dimensional representation of illumination we use principal component analysis (PCA) to reduce the dimensionality of the HDR environment images. We preprocess each image by resizing it to $64 \times 128$ and applying $\log(1 + I)$ where $I$ is the HDR image. The latter of the preprocessing steps is necessary as the HDR images can take a wide range of values, which would be difficult to model using PCA directly. In order to augment the training set we apply 5 random rotations to each image, resulting in a total of 1620 images used to fit the PCA model. The model reduces the HDR image dimensionality to 50 while explaining 80\% of the variance of the training data.



\subsection{Rendering}

To create a scene to be rendered each of the above mentioned assets (shape, texture, hair, and  illumination) is sampled independently.

\begin{figure}[t]
    \centering
    \includegraphics[width=\textwidth]{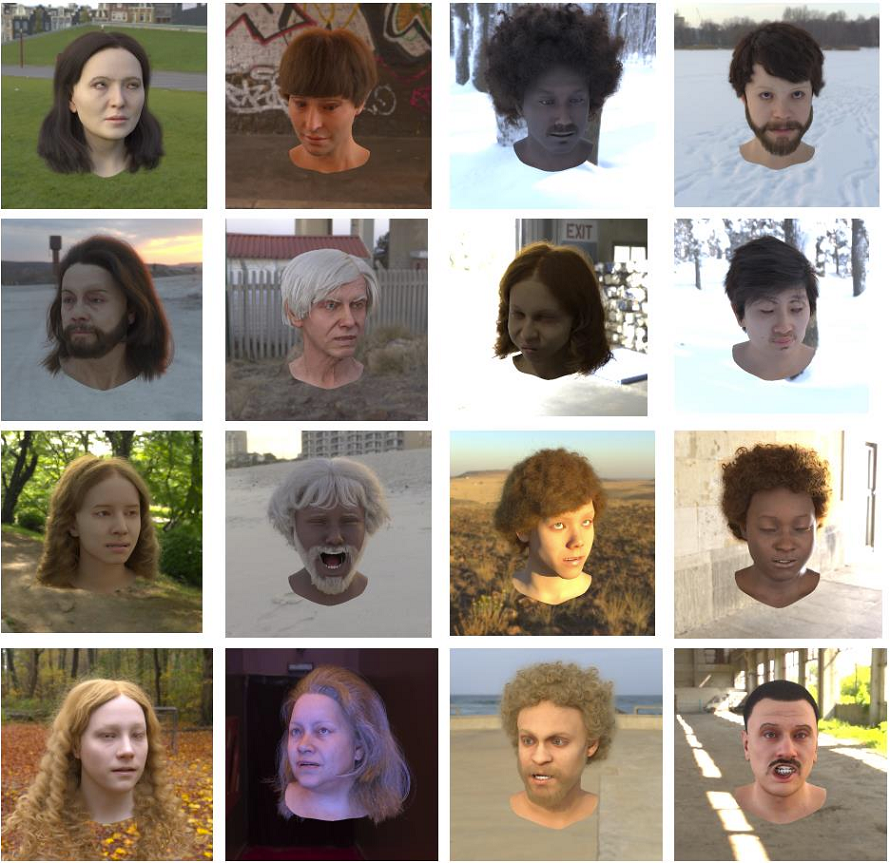}
    \caption{Examples of face image renders with randomly sampling shape, texture, expression, illumination, and hair. Note the diversity of generated images.}
    \label{fig:samples}
\end{figure}

The scene is then rendered using a frontal facing camera using Cycles renderer from Blender.
We render $1024\times1024$ images using 256 samples per pixel.
Examples of images rendered using our framework can be seen in Figure \ref{fig:samples}

\clearpage
%
%
\bibliographystyle{splncs04}
\bibliography{references}
\end{document}